# Fault Monitoring in Passive Optical Networks using Machine Learning Techniques

Khouloud Abdelli[1], Carsten Tropschug[2], Helmut Griesser[3], and Stephan Pachnicke[4]

1 *Nokia Bell Labs, Germany*
2 *ADVA Optical Networking SE, Germany*
3 *ADVA Network Security GmbH, Germany*
4 *Christian-Albrechts-Universität zu Kiel, Kaiserstr. 2, 24143 Kiel, Germany*
e-mail: Khouloud.Abdelli@nokia.com

**ABSTRACT**
Passive optical network (PON) systems are vulnerable to a variety of failures, including fiber cuts and optical network unit (ONU) transmitter/receiver failures. Any service interruption caused by a fiber cut can result in huge financial losses for service providers or operators. Identifying the faulty ONU becomes difficult in the case of nearly equidistant branch terminations because the reflections from the branches overlap, making it difficult to distinguish the faulty branch given the global backscattering signal. With increasing network size, the complexity of fault monitoring in PON systems increases, resulting in less reliable monitoring. To address these challenges, we propose in this paper various machine learning (ML) approaches for fault monitoring in PON systems, and we validate them using experimental optical time domain reflectometry (OTDR) data.

## 1. INTRODUCTION

Passive optical networks (PONs) have gained popularity as a broadband fiber access network solution due to their service transparency, cost effectiveness, and scalability among other benefits [1]. Because of their high capacity and extensive coverage, PON systems are becoming increasingly vulnerable to a variety of failures including failures in the optical distribution network (e.g., fiber cuts), optical network unit (ONU) transmitter/receiver failures, and dirty/cut/bent patch cords or connectors. Such failures, particularly fiber cuts, can result in network disruption and massive financial losses for service providers or operators. Fault detection in PONs requires complex manual intervention, as well as extensive expert knowledge and probing time until a failure is identified, located, and repaired, resulting in increased OPEX and customer dissatisfaction. As a result, implementing an accurate and efficient fault monitoring scheme in PON systems is extremely beneficial to reduce maintenance costs, to minimize downtime, and to improve service quality.

Optical time domain reflectometry (OTDR), a technique based on Rayleigh backscattering, has primarily been used to monitor optical fiber networks. However, using OTDR to monitor PON systems can be challenging because the backscattered signals from each branch are added together, making it difficult to distinguish between the individual branches' backward signals [2]. Event analysis becomes more difficult in the case of (almost) equidistant branch terminations because the reflected signals from the same length branches overlap and add up. Furthermore, the high loss of the optical splitters at the remote node causes a significant reduction in the backscattered signal, which may have an adverse effect on the event analysis. One proposed solution to address the aforementioned challenges is to use a tunable OTDR in conjunction with wavelength multiplexers to allow for a dedicated monitoring wavelength for each branch [3]. Such a solution, however, is prohibitively expensive due to the high cost of a tunable OTDR instrument, and its scalability is limited due to practical limitations and poor spectrum efficiency. Installing optical reference reflectors at the ends of each branch to check its integrity is another simple and effective solution [2]. However, this approach necessitates varying branch lengths, which limits its applicability to real-world installed networks. Recently, machine learning (ML)-based approaches have demonstrated great promise for improving fault monitoring in PON systems by extracting insights from OTDR monitoring data without the intervention of trained personnel or the installation of additional network infrastructure equipment [4]. In this paper, we propose and discuss different ML approaches for faulty branch or ONU identification in PON systems with similar or close branch lengths by leveraging monitoring data obtained from reflectors installed at each branch's end. The efficiency of each proposed method is validated using experimental OTDR data derived from a PON system.

## 2. A Network-dependent Approach

Such approach is trained on data from a PON system (i.e., long OTDR sequences including the reflections from all the branches). By learning the patterns of various faulty scenarios, including cases of a single faulty branch or multiple faults occurring at multiple fiber strands, the ML approach can identify the faulty branch identifier(s).

### 2.1 Experimental Data

The experimental setup shown in Fig. 1 is carried out to record OTDR traces incorporating faulty branch and normal operation conditions. A real passive optical network is reproduced by adopting a cascade of optical splitters leading to a splitting ratio of 1:128. The length difference between two close branches ranges from 2 m to 6 m, resulting in samples that include reflected pulses that may overlap. To model the faulty branch conditions, fixed optical attenuators with different attenuation settings (3 dB ... 8 dB) are used. For the sake of simplicity, it is assumed that multiple faulty branches cannot occur simultaneously. Therefore, for the data collection, the attenuator is inserted only at one branch at a time while keeping the other branches normal without placing any attenuators to produce "single faulty branch pattern" samples. Whereas the "normal pattern" samples are derived from the experimental setup that was performed without the addition of any attenuator. The OTDR configuration parameters, namely the pulse width, wavelength, and sampling time, are set to 10 ns, 1650 nm, and 2 ns, respectively. The laser power is adjusted from 0 to 16 dBm. The time required to collect and average the OTDR traces ranges from 2 ms to 2 s. From each sample (faulty or normal), a sequence of length 280 containing the reflections of all the branches is extracted. Because the range of considered attenuation values from 3 dB to 8 dB is narrow, and thus the resulting data is not diverse enough, the height of the faulty branch's reflection is artificially reduced, while the heights of the other branches' reflections remain as they are in normal operation. Given that the generated data is not very noisy, additive white Gaussian noise is added to each sequence in such a way that the peak to noise ratio (PNR) (defined as the ratio of the height of the reflection to the standard deviation of the noise) ranges between 5 and 30 dB (the noise functions as a form of regularization to enhance the robustness of the model). In total, a dataset composed of 44,460 samples (~4940 examples for each faulty branch and normal condition) is built, normalized, and divided into a training (60%), a validation (20%) and a test dataset (20%).

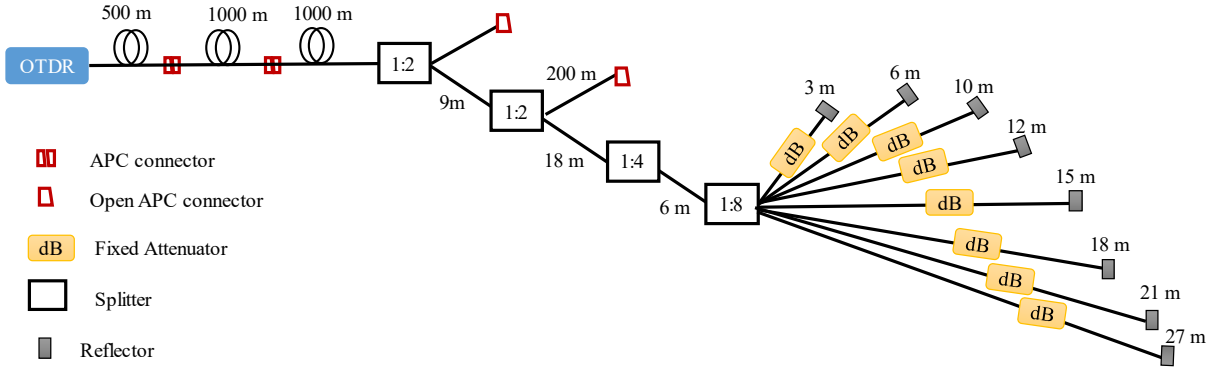

*Fig. 1: Experimental setup for generating faulty branch data in a passive optical network.*

### 2.2 Model Architecture

The architecture of the proposed ML model for faulty branch identification is illustrated in Fig. 2. The ML approach takes as input an OTDR sequence of length 280 incorporating the reflections of all the branches, and outputs the identifier of the faulty branch (0: normal (no faulty branch), $i$: branch $i$ faulty $\forall\ i\ \epsilon\ \{1, 2, 3, 4, 5, 6, 7, 8\}$). The input is fed to the hidden layers composed of three gated recurrent network (GRU) [5] layers that extract the relevant features underlying the patterns for normal and faulty branch conditions. The learnt features are then transferred to the output layer accompanied with a Softmax activation function which outputs the probability distribution over the different investigated classes. The model is trained by minimizing the categorical crossentropy (i.e., the cost or loss function) by adopting the Adaptive moment estimation (Adam) optimizer.

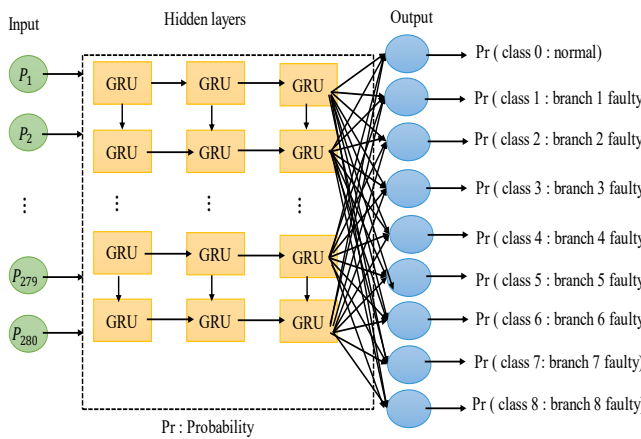

*Fig. 2: Architecture of the network-dependent ML model.*     *Fig. 3: The confusion matrix achieved by the model.*

### 2.3 Performance Evaluation & Discussion

The performance of the ML model is evaluated using an unseen noisy test dataset. The confusion matrix shown in Fig. 3 shows that the ML model accurately classifies the different faulty branch and normal conditions by achieving an average diagnostic accuracy of 97.45%. The model's misclassification rates can be explained by the fact that under low SNR conditions, the patterns of the normal and faulty branch conditions may look similar due to noise overwhelming the signal, making it difficult for the model to distinguish between the different patterns. To test the robustness of the ML model, the performance of the model is evaluated given unseen experimental data generated for a longer feeder fiber whereby a fiber break for a random branch is reproduced one at a time. A variable optical attenuator (VOA) located after the first 1km fiber is used to simulate the effect of a longer feeder fiber. The attenuation of the VOA is varied from 2 dB to 16 dB. The laser power and averaging time are set to low values to produce very noisy data. Tested with the generated data, the ML model achieves a good diagnostic accuracy of 93.4%. As demonstrated, the proposed ML model performs well, when tested with data derived from the same PON system used for training. However, if the PON topology changes (for example, by removing one of the ONUs or changing the length difference between ONUs), the ML model's performance may suffer because the learnt patterns modelling normal behaviour and the various faulty branch scenarios are altered. As a result, for each relevant network change, the ML model must be re-trained or fine-tuned, and its architecture must be adjusted accordingly.

## 3. Generic Approaches

The generic approaches are trained independently of a specific PON network topology and can work for different PON systems without having to be re-trained when the network changes. They extract insights from monitoring data composed of various smaller OTDR sequences with no more than two reflections to capture the relevant patterns that one or two reflections may exhibit under different faulty and normal conditions, and thus identify and localize the faulty branch(es). Figure 4 illustrates two different generic approaches [6, 7] that can be adopted for PON fault monitoring. PON systems with reflectors installed at each fiber branch are periodically monitored with OTDR. The recorded OTDR traces are collected and stored either locally or at a centralized database (e.g., in a software defined networking (SDN) controller). The OTDR signals are then normalized and split into fixed-length sequences with the split starting point following the optical splitter position. The length of the sequence is determined by several factors, including the pulse width, sampling rate, and the maximum and minimum length differences between neighbouring branches, and is chosen to be short to ensure that more than two reflections occur within a sequence is extremely unlikely. The processed data is then fed into the generic approaches for inference and prediction. The first method outputs simultaneously the reflection type (no reflection, one reflection, two reflections), the reflection event location(s), and reflection level(s) (the height(s) of the reflection(s)). The difference between the output and initial reflection levels is compared to a predefined threshold to determine the integrity of a fiber branch. Please note that a reference trace (i.e., an OTDR measurement performed when the PON system is deployed or when the network topology is changed) is used to calculate the initial reflection level(s). Given the output reflection event location(s), the faulty ONU(s) can then be identified and pinpointed. The second approach outputs the event type ($C_1$: two reflections where the first reflection is faulty, $C_2$: two reflections where the second reflection is faulty, $C_3$: two reflections both faulty, $C_5$: one reflection faulty) as well as normal conditions ($C_0$: two reflections both normal, $C_4$: one reflection normal, $C_6$: no reflection) and its location(s). The outcomes are then compared to the expected outputs in case of normal operation (extracted from the reference trace). The faulty branch can be identified based on the difference between predicted and expected outputs. Given the output location(s), the faulty ONU(s) can be localized.

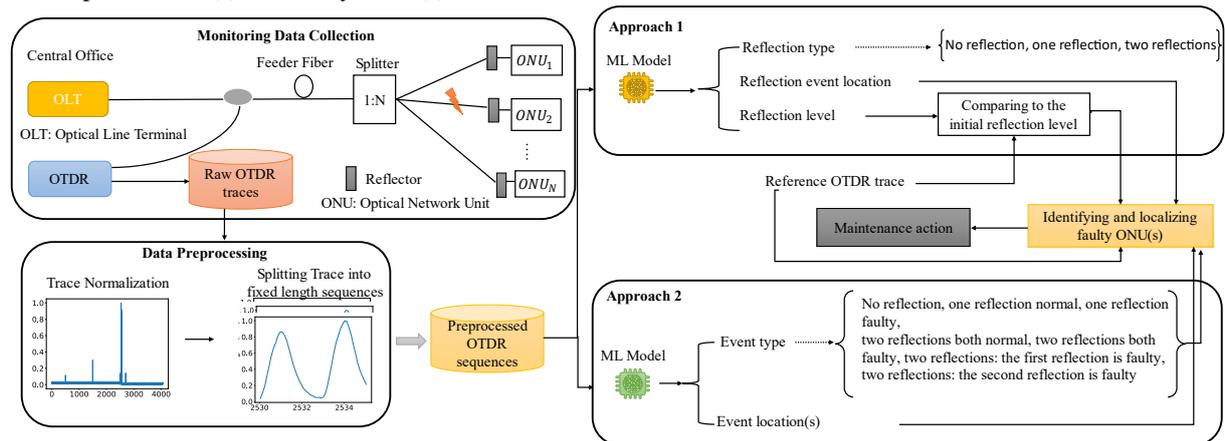

*Fig. 4: Proposed generic approaches for fault monitoring in PON systems.*

### 3.1 Validation Data

To validate the proposed generic approaches, the experimental setup shown in Fig. 1 is slightly modified by only putting the attenuators in the 3 m, 10 m, 12 m and 18 m branches. The placement of the attenuators is carefully chosen to get the sequences of all patterns representing the different investigated classes ($C_i \ \forall i \ \epsilon \{1…6\}$). For producing samples incorporating patterns for the class $C_0$, OTDR signals recorded under normal condition (OTDR traces generated using the experimental setup shown in Fig. 1 while removing all the attenuators) are adopted. The generated OTDR traces are then normalized and split into small sequences of length 30. To generate noisy data, the same principle of adding Gaussian noise is used. In total, a data set composed of 167,020 samples is built, and randomly split into a training (60%), a validation (20%) and a test dataset (20%).

## 3.2 Model Structures

Each ML method is made up of a long short-term memory (LSTM) layer with 16 cells that extracts the relevant features from the input, followed by task-specific layers that output the ML model's results based on the approach. For the first method, three task-specific layers of 16, 32, and 16 neurons output the reflection type, position of the reflection(s), and reflection level(s) respectively. In the second approach, there are two task-specific layers with 16 and 32 neurons that output the event type and event location, respectively (s). Each ML approach's total loss is calculated as the weighted sum of the individual task losses.

## 3.3 Results and Discussion

Figure 5 depicts the performance of the first approach. Fig. 5 (a) shows that the ML model classifies the various types of the investigated classes with a diagnostic accuracy greater than 95.9%. The "two reflections" and "one reflection" classes are rarely misclassified, especially when one of the reflections has vanished. As shown in Fig. 5 (b), the ML model accurately pinpoints the reflection(s) and thus the ONU(s) with very small position errors. The histogram in Fig. 5 (c) shows that the ML model accurately predicts the normalized reflection level(s) with small errors, implying that the identification of the faulty ONU(s) could be accurate. Figure 6 shows the performance of the second method. The confusion matrix in Fig. 6 (a) demonstrates that the ML method accurately identifies the different investigated event types with an average diagnostic accuracy of 97%. The ML model accurately localizes the various events, as shown in Fig. 6 (b), by providing small position prediction errors.

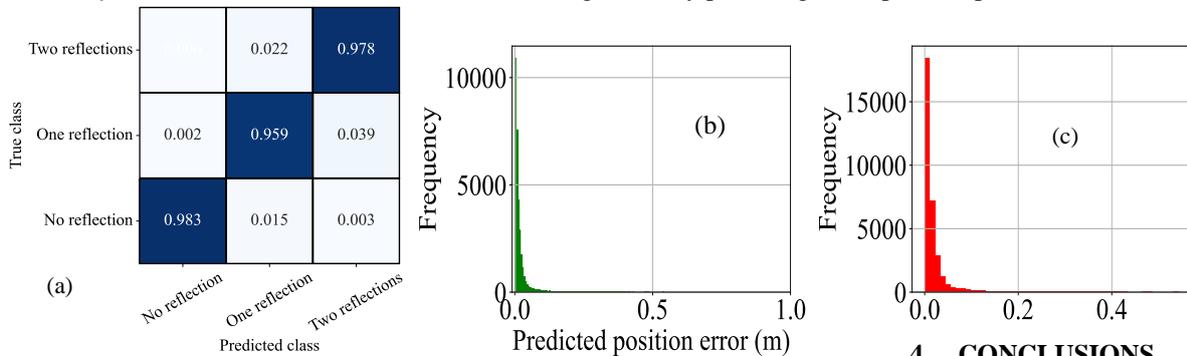

Fig. 5: Performance of the first generic approach: (a) Confusion matrix, (b) Histogram of the predicted position errors, and (c) Histogram of the predicted (normalized) reflection levels.

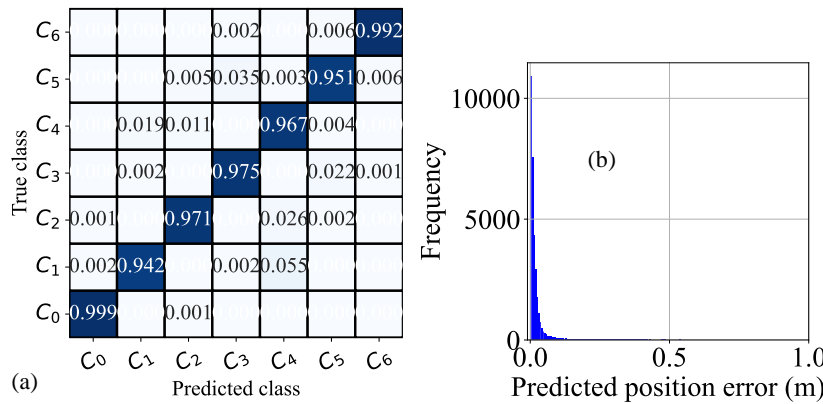

Fig. 6: Performance of the second generic method: (a) Confusion matrix, and (b) Histogram of the event position prediction errors.

## 4. CONCLUSIONS

Different ML approaches for faulty ONU(s) identification and localization in PONs have been proposed and experimentally validated using OTDR signals including normal and faulty branch conditions. The results prove that the proposed methods accurately identify and pinpoint the faulty branch(es), thereby improving fault monitoring in PON systems. Generic approaches outperform network-dependent methods because they do not need to be retrained for each network change.

**ACKNOWLEDGEMENTS**
This work was conducted while Khouloud Abdelli was working at ADVA Optical Networking SE and Kiel University.
This work has been performed in the framework of the CELTIC-NEXT project AI-NET-PROTECT (Project ID C2019/3-4).